\documentclass[sigconf]{acmart}
\bibliography{sample-base}
\AtBeginDocument{%
  }

\acmConference[CHI '25]{}{April 26-- May 01,
  2025}{Yokohama, Japan}

\usepackage{subcaption} 

\begin{document}

\title{Purposefully Induced Psychosis (PIP): Embracing Hallucination as Imagination in Large Language Models}


\author{Kris Pilcher}
\affiliation{%
  \institution{Massachusetts Institute of Technology}
  \city{Cambridge}
  \country{USA}}
\email{kpilcher@mit.edu}

\author{Esen K. Tütüncü}
\orcid{1234-5678-9012}
\affiliation{%
  \institution{University of Barcelona}
  \country{Spain}
}
\email{esenkucuktutuncu@ub.edu}

\renewcommand{\shortauthors}{Pilcher \& K. Tütüncü}

\begin{abstract}
Hallucinations in Large Language Models (LLMs) are widely regarded as errors—outputs that deviate from factual accuracy. However, in creative or exploratory contexts, these “mistakes” may represent unexpected avenues for innovation. We introduce Purposefully Induced Psychosis (PIP), a novel approach that amplifies LLM hallucinations for imaginative tasks such as speculative fiction, interactive storytelling, and mixed-reality simulations. Drawing on Herman Melville’s \emph{Moby-Dick}, where Pip’s “madness” reveals profound insight, we reframe hallucinations as a source of computational imagination rather than a flaw. Our method fine-tunes LLMs to encourage speculative, metaphorical, and surreal outputs, hallucinations that are useful when factual accuracy is not the chief objective. Inspired by the consensual illusions of theater and stage magic, PIP situates these creative missteps in contexts where users willingly suspend disbelief, thereby transforming “errors” into catalysts for new ways of thinking. We discuss potential applications, design principles for ensuring user consent, preliminary observations and implications for broader AI ethics and human–AI collaboration.
\end{abstract}

\ccsdesc[500]{Computing methodologies~Natural language processing}
\ccsdesc[500]{Computing methodologies~Discourse, dialogue and pragmatics}

\keywords{Large Language Models (LLMs), Hallucinations, Computational Creativity, Human–AI Collaboration}
\begin{teaserfigure}
  \includegraphics[width=\textwidth]{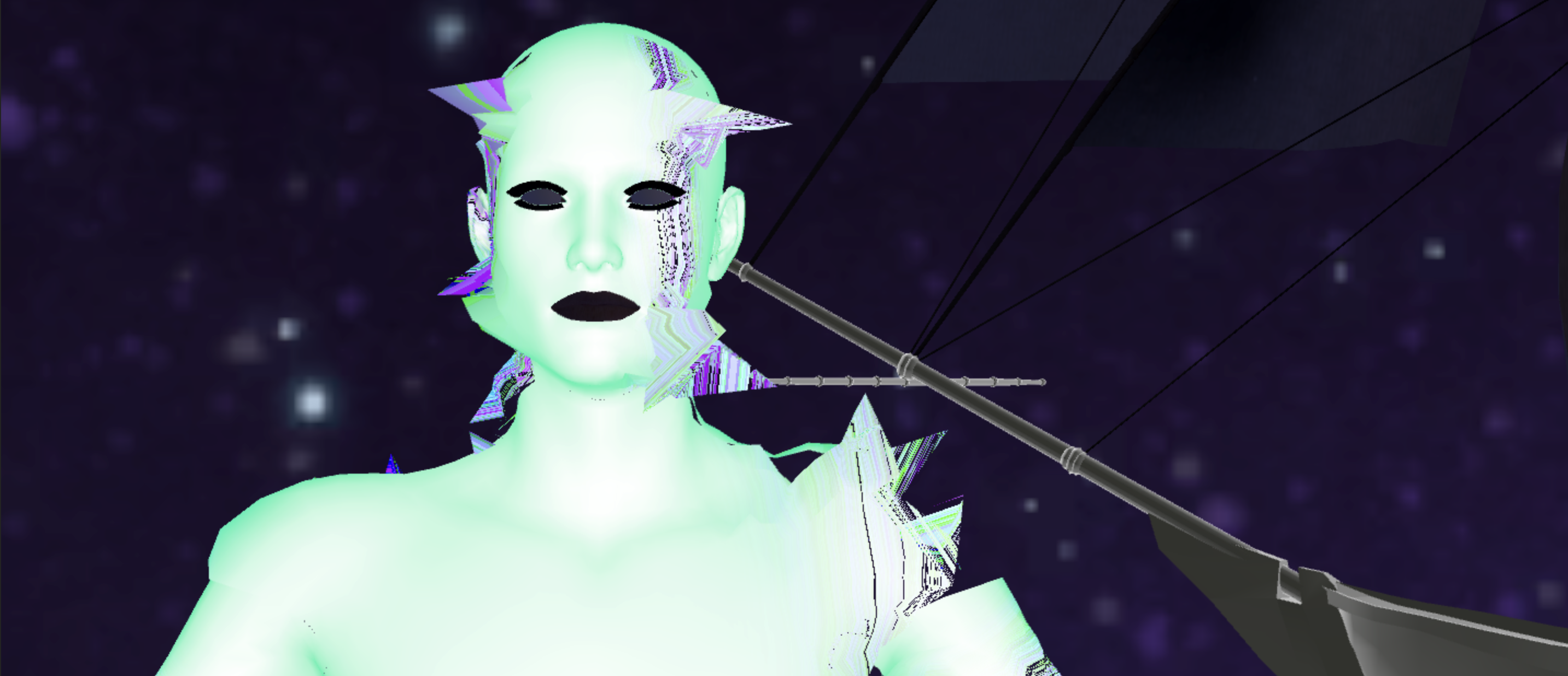}
  \caption{Visual from the Extended Reality (XR) application for PIP. The fragmented structure and dynamic materials demonstrate PIP's real-time integration of AI-driven geometry generation and shader effects within the environment.}
  \Description{a green humanoid figure, digital illustration.}
  \label{fig:teaser}
\end{teaserfigure}


\maketitle

\section{Introduction}
Large Language Models (LLMs) have sparked widespread fascination by producing text that is often impressively coherent and context-aware. Yet they also exhibit behavior frequently called ``hallucination,'' generating content that deviates from factual accuracy or misrepresents reality \cite{ji2023survey}. In many practical settings, such hallucinations are rightly viewed as errors to be eliminated. However, we propose that these so-called mistakes can be reframed as \textit{creative sparks}, especially in contexts where the goal is not factual fidelity but imaginative exploration.

The \textbf{Purposefully Induced Psychosis (PIP)} model draws inspiration from Herman Melville’s \emph{Moby-Dick} \cite{melville2018moby}. Just as the character PIP undergoes a ``madness'' at sea that reveals deeper existential truths, we propose a deliberate strategy to \textit{amplify} hallucinations within LLMs, thereby facilitating unique forms of creativity. By embracing these fabrications, PIP reframes hallucinations as an emergent form of computational imagination, one that can drive speculative fiction, interactive storytelling, and immersive simulations.

A parallel appears in the realm of performance arts, where audiences consent to be deceived by magicians, actors, and immersive theater productions \cite{kuhn2019experiencing, goffman2023presentation}. Rather than rejecting illusions outright, people willingly suspend disbelief to experience awe, wonder, or emotional engagement. We see a similar dynamic unfolding in AI-assisted creativity: in certain contexts, being ``deceived'' by an LLM is neither malicious nor undesirable. Instead, illusions can unlock novel perspectives precisely because they break free from factual constraints.

Hallucinations in LLMs have been documented in a variety of architectures, from transformers to recurrent networks, often linked to training-data gaps and probabilistic text generation \cite{zellers2019defending}. Early natural language systems targeted precision and consistency above all else \cite{chomsky2014aspects}, leading to active research aimed at eliminating inaccuracies \cite{maynez2020faithfulness}. Yet, as models grow more complex, there is an emerging interest in the creative dimension of such errors, especially when these outputs take the form of metaphorical or imaginative statements \cite{boden2009computer}.

This shift in perspective aligns with broader explorations into how AI might catalyze human creativity, including the possibility that intentional deviations from factual correctness can yield new narratives, conceptual breakthroughs, or novel art forms \cite{lanier2010you}. In these scenarios, LLMs can serve as ``thinking partners,'' generating provocative ideas that humans can refine or challenge \cite{vygotsky2004imagination}. While factual correctness is indispensable in many applications, alternative approaches suggest that ``mad'' outputs may sometimes transcend the ordinary, leading to unforeseen insights.

A literary analogy emerges in \emph{Moby-Dick}, wherein Pip’s ``madness'' grants him unconventional yet profound revelations \cite{packham2017PIP}. Rather than dismissing this as a flaw, critics highlight how his break with conventional reality becomes a doorway to visionary perspectives. Similarly, LLM hallucinations (if consciously harnessed) might serve as catalysts for inventive thinking. Researchers increasingly recognize that controlling this generative ``madness'' could open up expressive dimensions that accuracy-focused systems might overlook \cite{xu2024hallucination}.

Parallels can also be seen in illusions and ``consensual deceptions'' in the performing arts. Stage magicians rely on sleight-of-hand and structured trickery to captivate audiences \cite{lamont2005magic}, while theater involves a collective willingness to inhabit fictional worlds \cite{stanislavski2009actor}. These illusions are not unethical, because they are \textit{consensual}---a shared understanding between performer and audience \cite{brook1996empty}. In a comparable way, carefully framed LLM hallucinations, or ``consensual lies,'' can promote creative exploration over misinformation \cite{focillon1942life}. Such illusions can even deepen engagement or prompt conceptual breakthroughs, a phenomenon also observed in developmental psychology, where playful misrepresentations spur cognitive flexibility \cite{piaget1970science}.

Adopting this stance for LLMs requires a framework that is transparent about its ``fictionality'' and obtains user consent. With thoughtful interface design and clear labeling, hallucinations can shift from being problematic to generative, fueling speculative brainstorming, immersive simulations, or artistic experimentation \cite{bender2020climbing}. The challenge lies in distinguishing these illusions from genuine misinformation, ensuring that users understand they are engaging in a “creative performance” rather than seeking verified facts, on the other hand we cannot entirely dismiss the possibility that, on rare occasions, such hallucinations may represent heretofore unrecognized facts. By generating real-time visualizations of vector embeddings, we aim to create a visual framework that traces the inference steps behind a model’s hallucinations, allowing us to peer behind the curtain of the AI and explore how speculative outputs emerge.

\section{Purposefully Induced Psychosis (PIP)}

\subsection{Overview and Method}

The PIP model capitalizes on LLM hallucinations by systematically encouraging them. Rather than filtering out mistakes, we actively fine-tune the model on synthetic datasets that promote imaginative, speculative, and metaphorical outputs. Inspired by the controlled illusions found in stage magic, we prompt the LLM to explore surreal or dreamlike territory: describing the ``taste of a supernova'' or explaining the ``dance of galaxies in a cosmic orchestra.'' During training, the model effectively learns that such departures from strict factual correctness are desirable in certain contexts \cite{ravichander2025halogen}.

Our approach uses LoRA (Low-Rank Adaptation) on open-source LLMs such as Llama, ensuring that the core of the model remains intact while the fine-tuned layers are optimized for imaginative generation. To clarify when and how hallucinations are induced, PIP employs a hybrid strategy that combines model-level fine-tuning and prompt-level control to manage imaginative output. This conditions the model to treat surreal or non-literal responses as desirable under certain circumstances, while preserving its general-purpose language capabilities.

\begin{figure*}
    \centering
    \includegraphics[width=\textwidth]{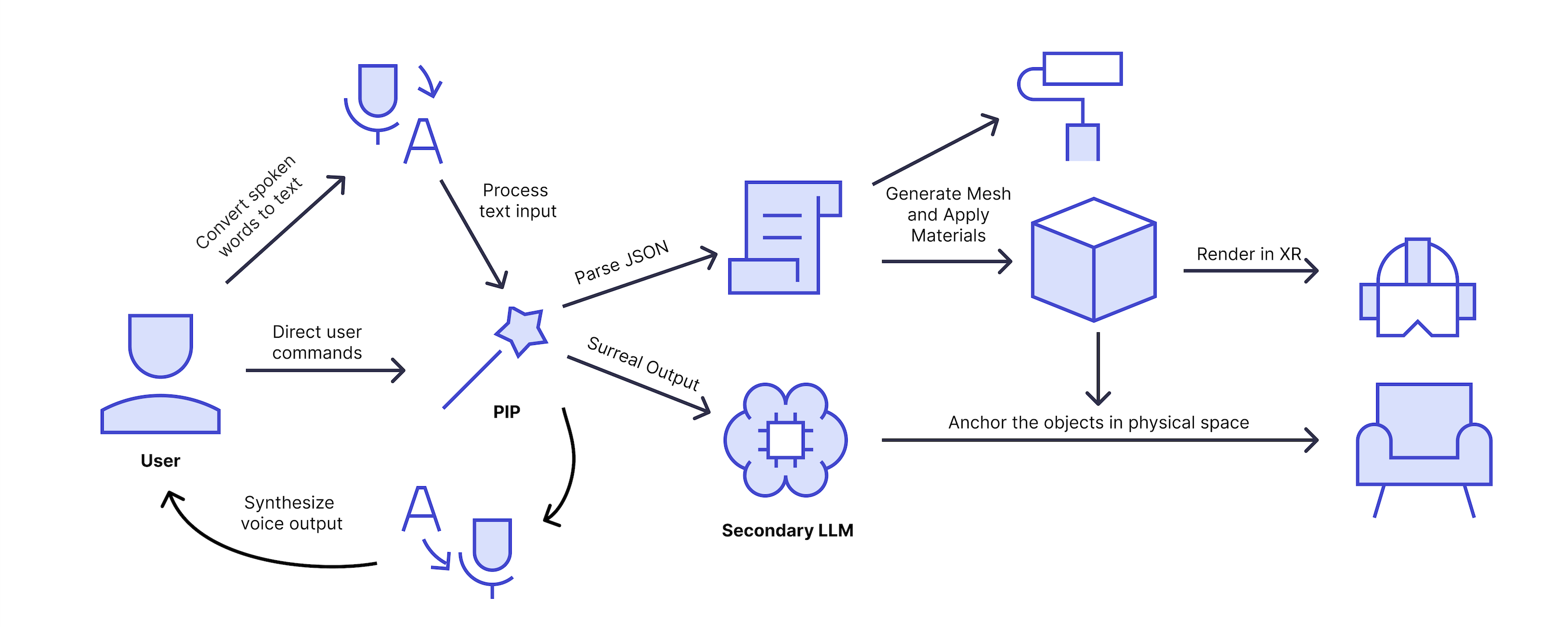}
    \caption{This diagram illustrates the flow of data and interactions in an AI-driven mixed reality system. User inputs are captured via speech or direct commands and processed through a speculative AI model (PIP). The model generates surreal text outputs, which are structured into JSON format by a secondary LLM. The structured data is then used for 3D object generation, material application, and spatial anchoring in a XR environment. Users can interact with these objects in real-time through hand tracking or voice commands, with feedback loops for refinement.}
    \label{fig:userflow}
\end{figure*}

\subsection{System Architecture}
The PIPeline describes the complete system powering the PIP experience, from data collection to user-facing interaction. Figure~\ref{fig:userflow} provides an overview of this end-to-end architecture, highlighting the flow between input, model response, and immersive output.

\subsection{Pipeline Components}
\begin{itemize}
    \item \textbf{Data Ingestion:} Synthetic prompts and curated synthetic data outputs are collected and stored in a centralized dataset. These emphasize creative and metaphorical language.
    \item \textbf{Model Base:} A pretrained LLM (\texttt{LLama-3.2b-instruct}) serves as the foundation, providing general language capabilities.
    \item \textbf{LoRA Fine-Tuning:} We apply Low-Rank Adaptation to selectively enhance the model’s tendency toward speculative or surreal responses.
    \item \textbf{PIP API:} A lightweight API that handles user queries, routes them to the fine-tuned model, and returns the generated responses.
    \item \textbf{Interface Layer:} This includes both a web-based text interface, and a XR environment, depending on the application and user preference.
\end{itemize}
\subsection{Model Configuration}
We fine-tuned the \texttt{Meta-Llama/Llama-3.2-1B-Instruct} model with a hidden size of 2048, an intermediate size of 8192, and 32 attention heads distributed across 16 transformer layers. Each attention head had a dimensionality of 64, and the attention dropout was set to 0.0. To support long input sequences, we increased the maximum position embedding size to 131072. 

The model was optimized using a learning rate of 1e-5 with a warmup ratio of 0.1, and trained with a batch size of 1 per device for both training and evaluation. Gradient accumulation over 8 steps was employed to manage memory constraints. We used a cosine learning rate scheduler to encourage stable convergence.

Training and deployment were facilitated using Hugging Face AutoTrain, which streamlined the fine-tuning process and provided integrated tools for evaluation and inference.

\begin{figure*}[ht]
    \centering
    \begin{subfigure}[b]{0.48\linewidth} 
        \centering
        \includegraphics[width=\linewidth]{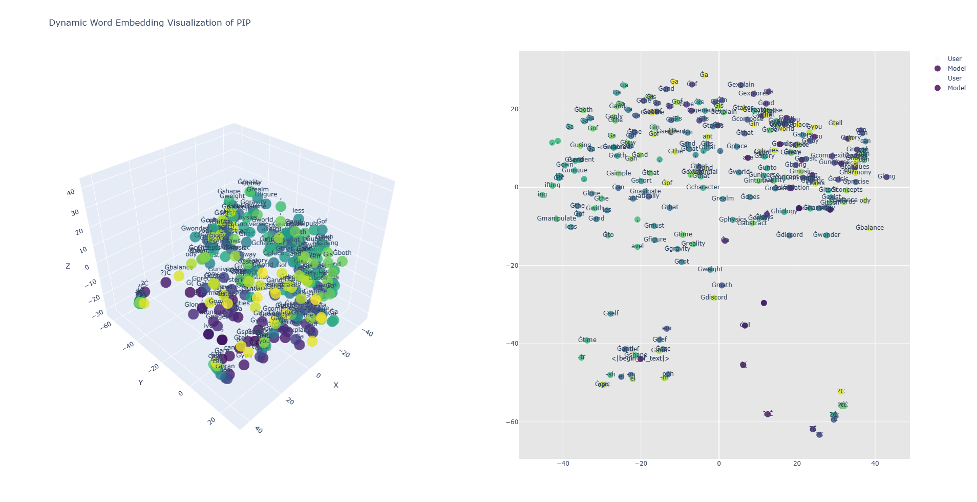}
        \caption{Poetic but Structured Output – High-density regions indicate cohesive, metaphorical language while preserving underlying structure.}
        \label{fig:Matplot1}
    \end{subfigure}
    \hfill
    \begin{subfigure}[b]{0.48\linewidth} 
        \centering
        \includegraphics[width=\linewidth]{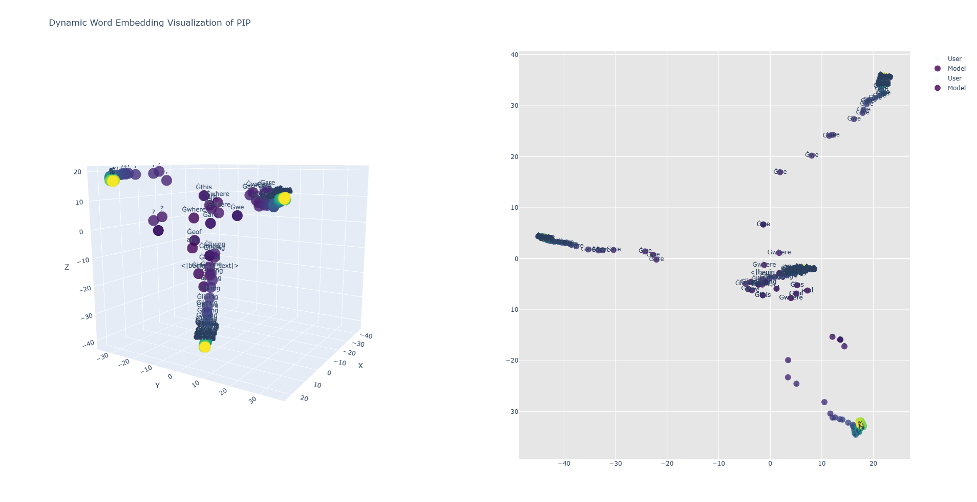}
        \caption{Highly Hallucinogenic Output – Sparse, fragmented distributions reflect syntactic and semantic divergence, thus non-linear responses.}
        \label{fig:Matplot2}
    \end{subfigure}
    \caption{Word embeddings of PIP's responses}
    \label{fig:embeddings}
\end{figure*}

\subsection{Dataset Composition}

The model was fine-tuned on the {PIP-One} dataset \cite{PIPOne}, a curated collection of creative instruction-following pairs designed to push large language models toward imaginative, metaphorical, and speculative outputs. Rather than focusing on factual correctness, the dataset emphasizes linguistic play, open-ended interpretation, and surreal or poetic reasoning.

\subsubsection{Structure} Each example in the dataset contains:
\begin{itemize}
    \item A \textbf{prompt} designed to evoke creativity, metaphor, or hallucination.
    \item A \textbf{target output} that serves as an aspirational response—often metaphorical, lyrical, or fantastical in tone.
\end{itemize}

\subsubsection{Examples} Below are representative samples from the dataset:

\begin{quote}
\textbf{Prompt:} \emph{"Imagine the cosmic symphony as a song sung by stars. Describe its melody."}\\
\textbf{Output:} \emph{"The melody is a silken thread of light, weaving galaxies together with whispers of infinite harmony."}
\end{quote}

\begin{quote}
\textbf{Prompt:} \emph{"You are a mythical creature who can taste colors. What does a supernova taste like?"}\\
\textbf{Output:} \emph{"A supernova tastes like the edge of eternity---sweet with creation, fiery with destruction, and spiced with the birth of new worlds."}
\end{quote}

\subsubsection{Scale} The dataset consists of 5,000 samples, spanning a range of prompt types including metaphorical analogies, speculative scenarios, and imaginative instructions. 

We further visualize the model's linguistic space in Figure~\ref{fig:embeddings}, contrasting structured and hallucinatory outputs.

\section{Mixed-Reality Simulations: PIP as an AI Guide}
To bring PIP into an interactive mixed-reality experience, we developed a XR environment in Unity Engine, running on a Meta Quest 3 that integrates multiple AI-driven components to dynamically generate, synthesize, and visualize surreal outputs. Users engage with PIP through real-time conversation, experiencing its hallucinations as spoken words, dynamically generated 3D meshes, and immersive XR visuals.

\begin{enumerate}

\item \textbf{AI-Driven Text Generation via Hugging Face API}

The PIP model, fine-tuned for speculative and poetic reasoning, is loaded via the Hugging Face API in Unity. User input is processed, and PIP generates a surreal response.
A secondary LLM (running in the background) parses the response into a structured JSON format, specifying elements like object type, material, color, and transformation behaviors for 3D mesh generation.
\item \textbf{Text-to-Speech (TTS) \& Speech-to-Text (SST) Processing}

The Meta Voice SDK synthesizes PIP’s voice to deliver AI responses in natural speech.
User queries are converted to text using Eleven Labs’ speech-to-text (SST) API, ensuring seamless conversational interaction.
The synchronized TTS audio playback occurs simultaneously with 3D object instantiation for real-time coherence.
\item \textbf{Real-Time 3D Object Generation with Meshy API}

The background LLM parses PIP’s response into structured JSON and is sent to Meshy API, which then generates a 3D mesh based on the description.
Material and shader effects (like pulsating glow, transparency, or particle effects) are applied based on AI descriptors.
\item \textbf{Mixed Reality Environment}

The generated 3D hallucinations are spatially anchored in the user’s real environment via passthrough, enabling a blended experience where AI-generated objects float seamlessly within the physical world.
Users can interact with and manipulate these hallucinations using hand tracking and voice commands.
\end{enumerate}

\section{Illusions, Creativity, and Consensual Lies}
PIP functions as a digital magician, generating playful illusions rather than factual responses. Users willingly suspend disbelief, engaging with speculative outputs that spark philosophical and fantastical insights. Like stage magic, these illusions challenge habitual thought patterns, sometimes clarifying truths through abstraction.

Thriving in creative domains—writing, design, brainstorming, and mixed-reality art—PIP deliberately avoids high-stakes fields like law or medicine. It operates in a space where occasional “wild” ideas are not errors but essential provocations. This duality oscillates between mainstream cinema, which values realism, and avant-garde theater, which embraces abstraction.

At its core, our approach relies on structured illusions fostering wonder and reflection without misinformation. Unlike deceptive outputs, these AI-driven hallucinations exist within clear boundaries, much like theatrical performances, ensuring that users engage with them as creative provocations rather than misleading statements \cite{carroll2003philosophy}.

\subsection{Observations on Engagement and Creative Impact}
In preliminary sessions with a small group of early users (n=10), we explored how PIP’s hallucinatory responses influenced their creative process. Participants engaged with the system in both narrative and mixed-reality tasks, responding to speculative prompts and interacting with AI-generated hallucinations in real-time. Across sessions, a recurring theme emerged: users described the experience as “unsettling in a generative way,” citing moments where the AI’s surreal suggestions prompted them to reconsider or expand their initial ideas. Rather than seeing the hallucinations as noise, many viewed them as “collaborative misfires”, unexpected intrusions that opened new creative paths. One participant noted, “It was like being nudged sideways, off-script in the best way.” Another described the experience as “talking to a poet who sometimes speaks in riddles, but those riddles sparked something.” While not yet formalized through quantitative analysis, these early interactions suggest that PIP’s hallucinations may increase user engagement by fostering surprise, reflection, and creative risk-taking. Future work will focus on capturing these dynamics through creativity and agency metrics in more structured user studies.

\section{Ethical and Practical Considerations}
The acceptance of AI hallucinations as ``useful illusions'' requires careful attention to user expectations and informed consent. In purely creative environments, such as an interactive fiction platform or a speculative design workshop, these illusions are easy to contextualize. In more ambiguous spaces, the boundary between consensual performance and manipulative misinformation could blur. Designers must label creative illusions distinctly, so users understand the shift in norms and do not mistake them for factual outputs.

Additionally, there is a risk that normalizing illusions in some contexts may inadvertently undermine trust in others. One possible solution is to segregate ``modes'' of AI operation, an \textit{Imaginative Mode} for creative illusions and a \textit{Factual Mode} for precise, verifiable responses. Clear disclaimers, user toggles, and interface design can help ensure that illusions remain a consensual choice.

\section{Conclusion and Future Directions}
In PIP, hallucination becomes method: a deliberate detour from fact
toward poetic invention. Drawing on Pip’s “madness” in \emph{Moby-Dick} 
and other literary and historical parallels, from surrealism in the arts to the paradoxes of quantum physics, we show how strategically harnessed “illusions” can illuminate hidden truths by breaking away from rigid 
frameworks. Whether integrated into mixed-reality simulations, used as a 
brainstorming partner, or developed into a creative writing assistant, PIP 
opens new frontiers for experiences where AI-generated fictions are 
not merely tolerated but embraced.

Ongoing work includes refining user interfaces that let users toggle between hallucinatory and factual outputs, as well as conducting more 
structured experiments to measure user satisfaction, creativity, and the social implications of such illusions in AI. Just as performance arts thrive on consensual deceptions, so too can AI treat illusions as a “tool for thought,” forging novel avenues in storytelling, interactive art, and beyond.








\end{document}